# Synthetic Data Generation and Automated Multidimensional Data Labeling for AI/ML in General and Circular Coordinates

Alice Williams
Department of Computer Science
Central Washington University
Ellensburg, WA, USA
Caden.Williams@cwu.edu

Boris Kovalerchuk
Department of Computer Science
Central Washington University
Ellensburg, WA, USA
Boris.Kovalerchuk@cwu.edu

*Abstract*—**Insufficient amounts of available training data is a critical challenge for both development and deployment of artificial intelligence and machine learning (AI/ML) models. This paper proposes a unified approach to both synthetic data generation (SDG) and automated data labeling (ADL) with a unified SDG-ADL algorithm. SDG-ADL uses multidimensional (n-D) representations of data visualized losslessly with General Line Coordinates (GLCs), relying on reversible GLC properties to visualize n-D data in multiple GLCs. This paper demonstrates use of the new Circular Coordinates in Static and Dynamic forms, used with Parallel Coordinates and Shifted Paired Coordinates, since each GLC exemplifies unique data properties, such as inter-attribute n-D distributions and outlier detection. The approach is interactively implemented in computer software with the Dynamic Coordinates Visualization system (DCVis). Results with real data are demonstrated in case studies, evaluating impact on classifiers.**

## I. INTRODUCTION

### A. Motivation

In many domains insufficient amounts of training data is a critical roadblock for the development and deployment of artificial intelligence and machine learning (AI/ML) models for multiple reasons: (1) to **train** ML models to **predict** from real-world data; (2) to **augment** under-represented cases of real data; and (3) to **anonymize** data for either privacy or legal issues [1-5].

For scenarios with **high-risk or high-stakes decisions** like that of cancer diagnostics, autonomous vehicle navigation, or market forecasting it is vital to generate extremely **high-quality** synthetic data to improve classifier performance and avoid catastrophic errors from erroneous classifications, then automating labeling saves valuable time of domain subject matter experts.

AI/ML models require an increased availability of high quality labeled data to improve training supervised models when data is highly multidimensional (n-D).

This paper addresses on-going critical challenges for both the tasks of **synthetic data generation** (**SDG**) and **automated data labeling** (**ADL**). Therefore, we propose a **unified SDG-ADL technology** to support AI/ML models, utilizing General Line Coordinates (GLCs) to losslessly visualize n-D data [7-11] with newly implemented Circular Coordinates (CC) originally defined in [7]. GLCs are reversible so we use multiple together.

### B. Overview of Existing Methods

There has been considerable progress made in deep learning Generative Adversarial Networks (GANs) methods for synthetic image generation [2]. A survey of current SDG techniques for both generation and evaluation from 2022 [1] listed several methods of SDG, see chart in Figure 1.

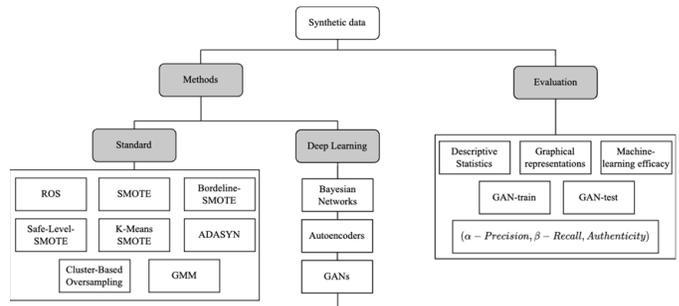

Figure 1. Existing methods for SDG and evaluation [1].

However, there is a remaining lack of generative methods for **synthetic tabular data**, which we focus on in this paper.

The SDG survey [1] identified major **challenges** of current state-of-the-art methods for generation of AI/ML data:

- Ensuring that synthetic data has the **same distribution** as that of the underlying real data.
- Ensuring that synthetic data generated by **black-box** methods, like deep learning GANs, can be **trusted.**

For instance, correlations between attributes and marginal distributions of both real and synthetic data can often appear deceptively similar. However, full n-D distributions can not be reliably identified with standard methods, due to the extremely small size of available real data relative to the size of the entire n-D feature space, worsening with increased dimensionality.

Real data, as seen in high-risk decision making scenarios, tends to exhibit this higher dimensionality, with an added need for quick generation and evaluation. Thus, **fundamentally new methods** are required for both the generation and evaluation of synthetic data, including for that of high-quality tabular ML/AI data.

## C. Data "Blindness" Challenges

What causes mismatches in distributions of n-D synthetic data from real data? Why is it difficult to interpret and explain SDG methods? A significant reason of both challenges is our inherent "blindness" to the distributions of data in highly n-D feature spaces. For instance, a most popular Synthetic Minority Over-sampling Technique (SMOTE) [15] uses *any training case of the minority class* to generate synthetic data. However, each training case influences the resultant accuracy of ML/AI models differently. Some cases are significantly more important to the classifiable features, often impacting features of other cases within the n-D feature space. SMOTE, "has been shown to yield poorly calibrated models, with an overestimated probability to belong to the minority class" [6] exhibiting this distributional "blindness". Selecting best real data cases as prototypes for SDG is critical.

More advanced SDG methods select real cases by importance determined from only **partial knowledge** of the real n-D data distribution. While these advanced methods are widely accepted [1], these methods have a fundamental issue of limited partial knowledge of real data distributions which are **not sufficient** for reliably selecting real data cases with the right methods to produce representative cases from. It is also a significant problem for ADL that real cases to be labeled have different impacts on the overall model accuracy.

As a result, current algorithms are inherently **"blind"** to real data distributions. Therefore, current methods are **heavily heuristic** and do not ensure that the results with extended synthetic data will truly produce more reliable models than without added synthetic data. Thus, **blindness** of n-D data distributions can lead to catastrophic failure.

## II. PROPOSED APPROACH

### A. Lossless Visualization of Multidimensional Data

Commonly used visual methods to evaluate the quality of synthetic data are **one-dimensional visualization** tools like box plots, histograms, and violin plots [1]. However, these methods can not adequately represent n-D AI/ML data **losslessly**, especially for our purpose of comparing highly n-D distributions of real and synthetic data. Commonly, dimensional reduction methods are used to handle highly n-D data, however, these are inherently **lossy**, preserving only a part of the overall n-D data properties and corrupting other properties. This is shown by the Johnson-Lindenstrauss lemma [7] which bounds the needed dimension $k$ to represent n-D data within a controlled loss of squared distance $\pm\epsilon$ error. The Johnson-Lindenstrauss lemma: Given $0 < \epsilon < 1$, a set $X$ of $m$ points in $\mathbb{R}^n$, and number $k > (8 \ln(m))/\epsilon^2$, there is a linear map $f: \mathbb{R}^n \to \mathbb{R}^k$ such that for all $u, v \in X$:

$$(1 - \epsilon)\|u - v\|^2 \leq \|f(u) - f(v)\|^2 \leq (1 + \epsilon)\|u - v\|^2$$

To overcome fundamental limitations of lossy visualization methods for n-D data we adapt **Visual Knowledge Discovery** (VKD) methodologies based on the **lossless and reversible** visualization methods of **GLCs** [7-11]. GLCs can **guide** both **SDG and ADL** to increase quality of the resulting model. In our approach both SDG and ADL **complement** each other. These processes identify areas where synthetic data can be *safely generated* and areas where unlabeled data can be *safely labeled*, and alternatively *avoided*.

Figure 2 losslessly visualizes the Iris 4-D dataset [12] where Setosa species are red, Versicolor are green, and Virginica are blue in different GLCs. Later in this paper we will illustrate the proposed methods with these data and their GLCs visualizations.

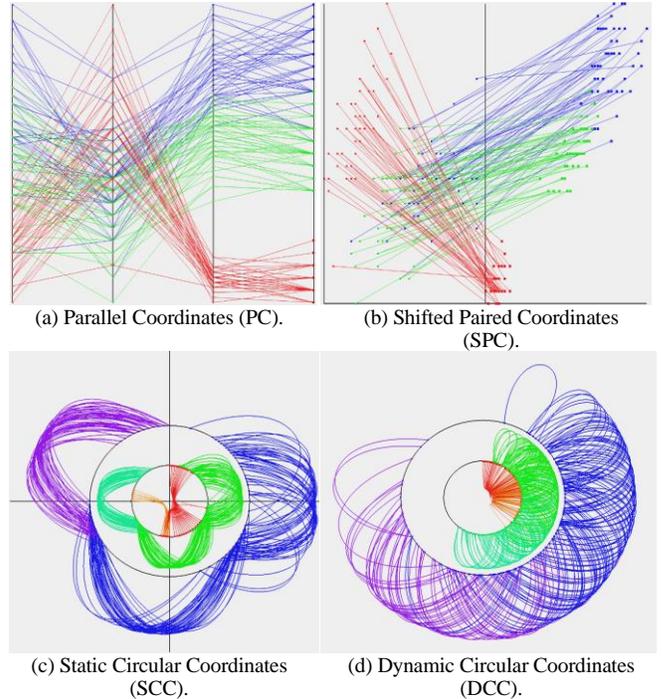

(a) Parallel Coordinates (PC).  (b) Shifted Paired Coordinates (SPC).

(c) Static Circular Coordinates (SCC).  (d) Dynamic Circular Coordinates (DCC).

Figure 2: GLC visualization methods for 4-D Iris data.

### B. Advantages of the Approach over Alternatives

One current SDG and evaluation method reviewed in [1] assumes that a given dataset $D$ is **split** into a train-set, $D_{train}$, and test-set, $D_{test}$. ML models trained on $D_{train}$, and on $D_{synth}$ (the synthetic data) are then compared and evaluated against, $D_{test}$. If the performance (accuracy, recall, precision, F1-score) of the models trained using $D_{train}$ is like those trained using $D_{synth}$, then, the conclusion is made that the synthetic data are likely to follow the **underlying data distribution** [1, 3, 4]. This is an **overoptimistic expectation**, which can not be taken as granted for high-risk tasks.

First, this expectation is applicable only for the data **split actually used**. It can be false for all other data splits. The number of possible splits is growing exponentially with the size of the dataset. Testing all possible splits is not feasible.

Second, it is not clear if a real dataset $D$ itself follows the underlying distribution of the **whole data**, which is never known. Thus, expected data consistency only holds between models produced from available real data and simulated data using just the specific split data used.

Third, the described split method is not applicable to tasks where we have an insufficient amount of training data for models, but where we want to **expand** the training data with synthetic data. This is the task that we consider in this paper.

A **modified method** splits not just the given real dataset, $D$ into a train-set, $D_{train}$, and test-set, $D_{test}$ but the entire *extended dataset* including **synthetic data**. The motivation is that an insufficient amount of training data disqualifies a model built only on $D_{train}$ to be an appropriate standard to evaluate the quality of models built on other data.

Now, the selection of only one or a few specific splits in k-fold cross validation can be misleading which can produce a drastically **inflated accuracy** for real data as shown in [9, 10].

The evaluation method from [1] discussed above tests the model on real data, for the model that was built on both real and synthetic data. Other test methods can give a more complete picture [5], where ***α*-Precision** measures the **similarity** between synthetic and real cases, ***β*-Recall** evaluates **diversity** of synthetic data relative to real data, and **authenticity** measures how well the model can **generalize**, therefore, not overfit to real data. While these measures are useful, they are not sufficient to ensure accuracy of models built with synthetic data. However, we can use these methods to **discard** low-quality cases [5].

### C. Proposed Visual SDG-ADL Algorithm

Our proposed SDG-ADL algorithm operates on all triplets of $\langle A_i, S_j, D \rangle$, where $A_i$ is a supervised ML classification algorithm, $S_j$ is a SDG algorithm, and $D$ is given real ML data. We are interested in finding the most efficient pair $\langle A_i, S_j \rangle$ for the given ML data $D$. The SDG-ADL algorithm steps are:

Step 1. *Visualize* n-D ML data $D$ in a selected GLC losslessly. (In this paper we use GLCs of PC, SPC, SCC, and DCC).
Step 2. *Find* and outline the *Most Pure* (MP) areas, pure being regions where only one label is assigned to the selected area.
Step 3. *Find* and outline the *Least Pure* (LP) areas.
Step 4. *Order* remaining areas by *Purity Levels* (PL),
Step 5. *Hide* absolutely pure areas to decrease occlusion.
Step 6. *Assign* Synthetic Data and Labeling to *Most Pure* areas.
Step 7. *Evaluate* Performance of SDG algorithm on all data.
Step 8. *Modify* step 6 repeating according to results from Step 7.
Step 9. *Assign* Synthetic Data and Labels to areas *outside* of the *Least Pure* area if Step 8 did not produce desired results.
Step 10. *Repeat* from Step 7 to Step 9 to improve results.
Step 11. If misclassification remains *repeat* from Step 1, select another GLC that shows data properties of remaining MP areas.

### D. Static Circular Coordinates (SCC)

Circular Coordinates were first introduced in [9]. Below we will call them **Static Circular Coordinates** (**SCC**) to contrast with a variant definition of Circular Coordinates that we will call Dynamic Circular Coordinates, defined later. SCC are constructed from tabular data, where each n-D point is a row labeled by its class. Numeric attributes are normalized using min-max normalization to the range of $[0, 1]$:

$$x' = (x - \min(x)) / (\max(x) - \min(x))$$

where max(x) and min(x) are minimum and maximum values of the attribute. Each coordinate occupies a section of the circumference of the circle, e.g., if data contains four coordinates, then four non-overlapping sections are created that cover all the circumference. Each individual attribute value $x_i$ of the n-D point $\mathbf{x} = (x_1, x_2, \ldots, x_n)$ is plotted as a vertex at the distance $x_i$ on the circumference from the start point of the segment assigned for this attribute. Vertices are connected by Beziér curves connecting each vertex to the next vertex where vertex $x_i$ connects to the next vertex $x_{i+1}$. Vertices with connected Beziér curves are uniquely colored by a class label.

Figure 2 demonstrates a single data class, Setosa, from Iris data visualized in SCC, shown in red with 50 cases. This single class of data has four components, giving us four vertices $x_i$ to each 4-D point $\mathbf{x} = (x_1, x_2, x_3, x_4)$ plotted with three inter-connecting Beziér curves of vertices $(x_1, x_2)$, $(x_2, x_3)$, and $(x_3, x_4)$. The circle is subdivided into four sectors by black axes.

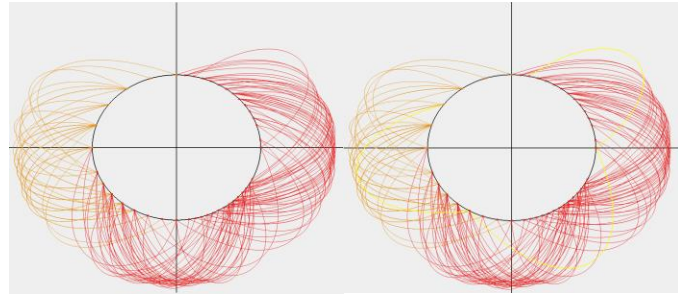

(a) SCC 4-D visualization.  (b) Selecting a single case in yellow.
Figure 3. SCC demonstrated with 4-D Iris data single Setosa data class (a) shown next to same visualization and data with a single case selected (b).

### E. Demonstrating steps 1-3 of SDG-ADL with GLCs

Figures 4 and 5 visualize the Iris data in both PC and SCC for discovering MP and LP areas with the heaviest inter-class case overlap, where cases are hardest to classify. These figures illustrate the SDG-ADL algorithm steps 1-3. Figure 4 shows MP and LP areas, with colored rectangles on the coordinates show absolutely pure MP areas. Yellow polylines highlight some overlap cases in LP area. Figure 5 shows SCC with MP areas denoted by sectors 1 and 3 and LP area is sector 2.

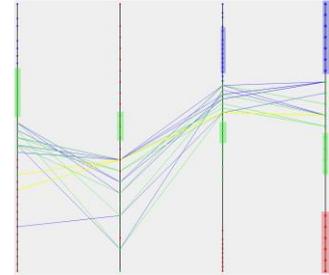

Figure 4. MP and LP areas of Iris 4-D data in PC. Rectangles on axes outline MP areas, cases in MP area have been hidden, simplifying observation of LP areas. Yellow polylines highlight some overlap cases in LP area.

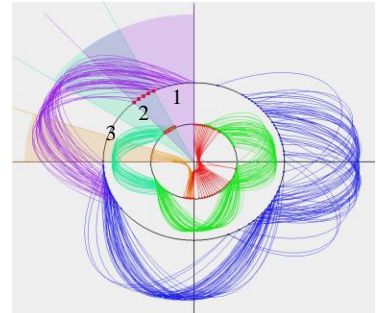

Figure 5. 4-D Iris data of three classes in SCC. MP areas are shown by segments 1, 3 and LP area are shown by segment 2. The yellow highlighted case is example of the case from the LP area.

Cases contained in the overlap LP area in Figure 5 require discovering more complex rules than rules based on a single coordinate used to find it. Visualization of the cases in this overlap area with Shifted Paired Coordinates defined in [7] shown in Figure 6 allows for this discovery. This visualization involves relations between *all four coordinates* and is visible in Figure 6, where more green lines (cases) slope downwards than blue lines. Here 8 green case lines slope *downwards (negative derivative)* and one green case line trends upwards *(positive derivative)*, while 3 case lines are horizontal (total 12 green lines). In contrast 4 blue case lines slope *upwards*, and 5 blue case lines trend *downwards*, while one case stays *horizontal (zero derivative)* with a total of 10 blue case lines. This leads to a classifying slope rule $R_s$:

*If a line slopes upwards (has a positive derivative), then it is in the blue class, else it is in the green class.*

Table 1. Confusion matrix for rule $R_s$.

|  | Predicted blue | Predicted green | precision |
|---|---|---|---|
| Actual blue | 4 | 6 | 4/10 (40%) |
| Actual green | 1 | 11 | 11/12 (91.67%) |
| Precision | 4/5 (80%) | 12/16 (75%) | **15/22 (68.18%)** |

Accuracy is 68.18% due to 15/22. See the confusion matrix (Table 1). We have a total of 7 misclassified cases out of 150 cases with total 95.33% accuracy of **all rules constructed completely visually without any complex machine learning models** used. Support Vector Machine (SVM) results with 4 misclassified cases, but with 59 support vectors [7, p. 121], which is not interpretable. In Figure 6 yellow circles mark two pairs of very similar green and blue lines of cases from two distinct labels which both go down. We keep them in the SPC overlap area. Classifying the remaining cases may require using additional complementary GLC(s).

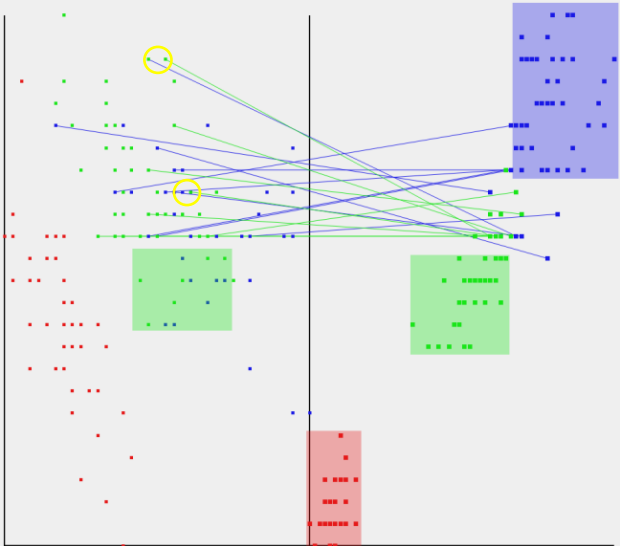

Figure 6. Visualization in SPC of the overlap cases after removing cases covered by first 5 rules from Figure 3 with coordinate X2 inverted.

### F. Dynamic Circular Coordinates (DCC)

Below we define Dynamic Circular Coordinates (DCC), DCC were first introduced under the influence of previously successful ideas of SCC, GLC-Linear [7], and Dynamic Elliptic Paired Coordinates [13]. In these GLC visualizations, the location of each attribute $x_i$ depends upon the location of the previous attribute $x_{i-1}$, going the attribute length out from the former point with the first $x_1$ positioned at the northernmost point on the axis going clockwise. Then, coefficients of the visualized linear discriminant function (LDF) variables can be scaled to visually tune the separation of data classes, benefiting from thorougly data labeling. In DCC, visualizing a n-D point $\pmb{x} = (x_1, x_2, ..., x_n)$ also is a sequence of Beziér curves like SCC. The first Beziér curve connects $x_1$ to $x_2$ and $x_2$ to $x_3$ continuing with curves to connect each point to the next. The difference is that the points themselves $x_i$ are dependent on each other after the first. In contrast, in SCC the locations of points $x_1, x_2, ..., x_n$ is fixed in the circle segments dedicated to each coordinate $X_1, X_2, ..., X_n$. Figure 7 shows the same 4-D Iris data for Setosa class as in Figure 3.

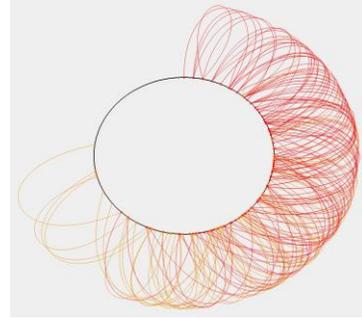

Figure 7. DCC demonstrated on 4-D Iris data with single Setosa class coefficients of [1, 1, 1, 1].

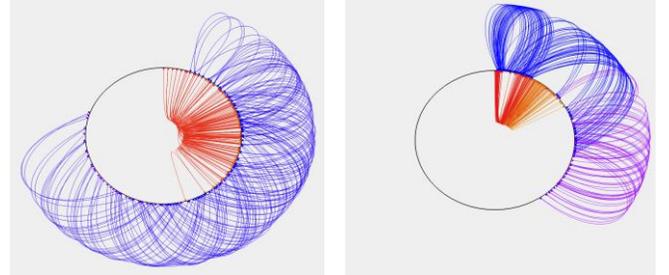

(a) coefficients of [1, 1, 1, 1].    (b) coefficients of [0.08, 0.43, 0.89, 0.52].

Figure 8. DCC with 4-D Iris data two class Setosa (red) and Versicolor (blue) with two unique sets of Linear Discriminant Function coefficients.

### III. CASE STUDIES

#### A. Benchmark generation and case study strategy

The following sections present case studies of generating synthetic data and classification models in different ways. To be able to compare them, we first need benchmarks, evaluated on only actual real data without any synthetic cases.

To benchmark our case studies, we first have performed a Monte-Carlo simulation of 1000 executions of training and validating with fourteen commonly used classifiers on the Iris data. The random split of data to training and validation was 90%:10% in 10-fold cross validation (CV). We repeated these values for execution count and cross validation folds in all case studies. The stability of the standard deviation (STD) was used as a criterion of accuracy robustness for the evaluated models. Classifiers evaluated with are: Decision Tree (DT), Random Forest (RF), Extra Trees (XT), K-Nearest Neighbors (KNN), Support Vector Machine (SVM), Linear Discriminant Analysis (LDA), Logistic Regression (LR), Ridge Regression (Ridge), Gaussian Naïve Bayes (NB), Multi-Layer Perceptron (MLP), Stochastic Gradient Descent (SGD), Gradient Boost (GB), AdaBoost (AB), Extreme Gradient Boosting (XB). All case studies were executed on the same consumer grade hardware.

Consider an example with all 150 cases of Iris data where 135 cases are randomly selected to be training data and 15 cases to be validation data in each 10-fold Cross Validation round. Next, we consider 250 cases, with 120 real cases and 130 synthetic cases. We want to know how successful models will be, which we generated on the benchmark data to classify all these 250 cases. We will call these data an *exploration data* (Exp. in tables) and accuracy of the models built in the benchmark study as *default classifier performance*. Benchmark results are first listed in Table 2.

Table 2. Iris benchmark tests on the classifier ensemble. Training data used is Iris, validation data is also Iris.

```
Model Performance over 1000 independent cycles with 10-Fold Cross-Validation
Training Dataset: C:/Users/Alice/Documents/GitHub/DCVis/datasets/fisher_iris.csv
Exploration Dataset: C:/Users/Alice/Documents/GitHub/DCVis/datasets/fisher_iris.csv
=================================================================================
Model  CV Mean Acc.  CV STD of Acc. Exp. Mean Acc. Exp. STD of Acc.Best AUC  Worst AUC
DT     0.94          0.01           1.00           0.00            1.00      1.00
RF     0.95          0.00           1.00           0.00            1.00      1.00
XT     0.95          0.00           1.00           0.00            1.00      1.00
KNN    0.95          0.00           0.97           0.00            1.00      1.00
SVM    0.96          0.00           0.97           0.00            1.00      1.00
LDA    0.97          0.00           0.98           0.00            1.00      1.00
LR     0.97          0.00           0.97           0.00            1.00      1.00
Ridge  0.83          0.00           0.85           0.00            0.96      0.96
NB     0.95          0.00           0.96           0.00            0.99      0.99
MLP    0.97          0.01           0.98           0.01            1.00      1.00
SGD    0.83          0.04           0.85           0.11            0.99      0.98
GB     0.95          0.00           1.00           0.00            1.00      1.00
AB     0.93          0.00           0.96           0.00            1.00      1.00
XB     0.94          0.00           1.00           0.00            1.00      1.00
=================================================================================
```

For these case studies we: (1) take balanced Iris data, (2) cut out some cases of a given class causing an imbalance, (3) generate synthetic data to rebalance, (4) compute accuracy of classification trained from synthetic data, and (5) compare accuracy with and without synthetic data to evaluate the SDG efficiency. We consider the following strategies of SDG:

(G1) Within the pure areas of cases of the same given class.
(G2) Outside of the pure areas of cases of the same class.
(G3) Randomly throughout the graph.

Our hypotheses for these SDG methods are that G1 will yield the same or an improved accuracy, G2 will decrease accuracy, and G3 will be unstable with accuracy going in both directions depending on the random synthetic data generated.

### A. Case Study I: Single Synthetic Case Strategy

This case study shows a significant advantage from lossless visualization of n-D data for generation of synthetic n-D data. Figure 9 shows a case marked with a yellow highlight. This case is an *outlier* of the red (Setosa) class on the second coordinate $X_2$ We added a single synthetic 4-D point slightly above that 4-D point to explore how sensitive classifier algorithms are toward such outliers when their number will increase.

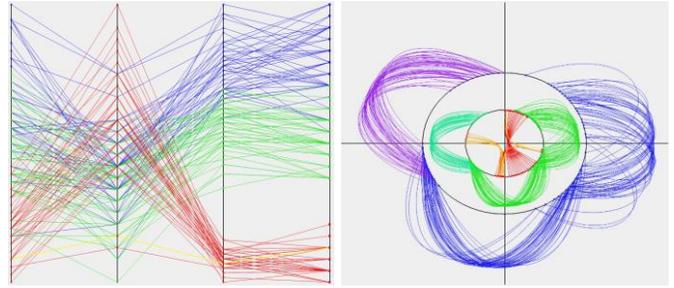

(a) Inserted case in PC.  (b) Inserted case in SCC.

Figure 9. Synthetic red case, highlighted in yellow, inserted into the Setosa class and placed directly above the real outlier case in $X_2$.

Table 3 shows the accuracy of classification of data after adding the synthesis 4-D point. It did not impact accuracy for most of the algorithms listed in Table 3, but for SGD algorithm it improved the accuracy by about 4.5%. This demonstrates robustness of classification algorithms regarding this outlier. The reason for this is the separation of the red (Setosa) class in both coordinates $X_3$ and $X_4$, which is visible in Figure 9. Without such visualization it would be difficult to determine the source of robustness for the resultant synthetic data.

Table 3. Accuracies for Case Study 1 (one synthetic Setosa case added).

```
Model Performance over 1000 independent cycles with 10-Fold Cross-Validation
Training Dataset: C:/Users/Alice/Documents/GitHub/DCVis/datasets/experiments/iris_setosa_synth1.csv
Exploration Dataset: C:/Users/Alice/Documents/GitHub/DCVis/datasets/fisher_iris.csv
=================================================================================
Model  CV Mean Acc.  CV STD of Acc. Exp. Mean Acc. Exp. STD of Acc.Best AUC  Worst AUC
DT     0.94          0.00           1.00           0.00            1.00      1.00
RF     0.95          0.00           1.00           0.00            1.00      1.00
XT     0.95          0.00           1.00           0.00            1.00      1.00
KNN    0.97          0.00           0.97           0.00            1.00      1.00
SVM    0.97          0.00           0.97           0.00            1.00      1.00
LDA    0.98          0.00           0.98           0.00            1.00      1.00
LR     0.97          0.00           0.97           0.00            1.00      1.00
Ridge  0.82          0.00           0.85           0.00            0.96      0.96
NB     0.95          0.00           0.96           0.00            0.99      0.99
MLP    0.97          0.01           0.98           0.01            1.00      1.00
SGD    0.83          0.04           0.85           0.11            0.99      0.98
GB     0.95          0.00           1.00           0.00            1.00      1.00
AB     0.94          0.00           0.96           0.00            1.00      1.00
XB     0.94          0.00           1.00           0.00            1.00      1.00
=================================================================================
```

Similar experiment with SCC is in Figure 10 and Table 4.

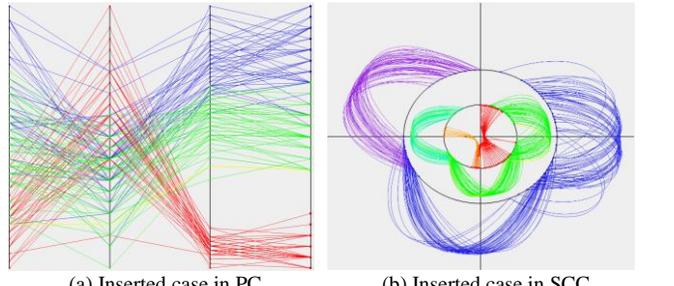

(a) Inserted case in PC.  (b) Inserted case in SCC.

Figure 10. Synthetic green case, highlighted in yellow, inserted into the Versicolor class and placed directly below last case in $X_1$.

Table 4. Results with one synthetic Versicolor case added.

```
Model Performance over 1000 independent cycles with 10-Fold Cross-Validation
Training Dataset: C:/Users/Alice/Documents/GitHub/DCVis/datasets/experiments/iris_versicolor_synth1.csv
Exploration Dataset: C:/Users/Alice/Documents/GitHub/DCVis/datasets/fisher_iris.csv
=================================================================================
Model  CV Mean Acc.  CV STD of Acc. Exp. Mean Acc. Exp. STD of Acc.Best AUC  Worst AUC
DT     0.95          0.00           1.00           0.00            1.00      1.00
RF     0.95          0.00           1.00           0.00            1.00      1.00
XT     0.95          0.00           1.00           0.00            1.00      1.00
KNN    0.97          0.00           0.97           0.00            1.00      1.00
SVM    0.95          0.00           0.97           0.00            1.00      1.00
LDA    0.98          0.00           0.98           0.00            1.00      1.00
LR     0.97          0.00           0.97           0.00            1.00      1.00
Ridge  0.84          0.00           0.86           0.00            0.96      0.96
NB     0.95          0.00           0.97           0.00            1.00      1.00
MLP    0.97          0.01           0.98           0.01            1.00      1.00
SGD    0.83          0.04           0.85           0.11            0.99      0.98
GB     0.95          0.00           1.00           0.00            1.00      1.00
AB     0.94          0.00           0.96           0.00            1.00      1.00
XB     0.95          0.00           1.00           0.00            1.00      1.00
=================================================================================
```

## B. Case Study II: Full Data Duplication Strategy

First we cloned all cases of Iris data and shifted them up by adding 0.1 to create synthetic cases of the same classes (see Figure 11). This had positive results shown in Table 3.

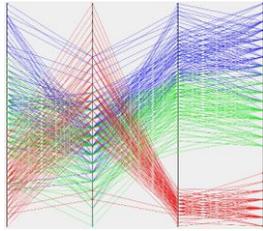

Figure 11. Iris with all cases duplicated and then shifted up by adding 0.1.

Table 3. Classifier models with almost perfect performance across the board.

```
=============================================================================
Model Performance over 1000 independent cycles with 10-Fold Cross-Validation
Training Dataset: C:/Users/Alice/Documents/GitHub/DCVis/datasets/experiments/test11.csv
Exploration Dataset: C:/Users/Alice/Documents/GitHub/DCVis/datasets/fisher_iris.csv
=============================================================================
Model   CV Mean Acc.   CV STD of Acc.  Exp. Mean Acc.  Exp. STD of Acc. Best AUC  Worst AUC
DT       0.96           0.00            1.00            0.00             1.00      1.00
RF       0.97           0.00            1.00            0.00             1.00      1.00
XT       0.97           0.00            1.00            0.00             1.00      1.00
KNN      0.96           0.00            0.96            0.00             1.00      1.00
SVM      0.97           0.00            0.97            0.00             1.00      1.00
LDA      0.98           0.00            0.98            0.00             1.00      1.00
LR       0.97           0.00            0.98            0.00             1.00      1.00
Ridge    0.84           0.00            0.86            0.00             0.97      0.97
NB       0.95           0.00            0.96            0.00             1.00      1.00
MLP      0.98           0.00            0.98            0.00             1.00      1.00
SGD      0.89           0.03            0.90            0.08             1.00      0.99
GB       0.97           0.00            1.00            0.00             1.00      1.00
AB       0.94           0.00            0.94            0.00             1.00      1.00
XB       0.96           0.00            1.00            0.00             1.00      1.00
=============================================================================
```

## C. Case Study III: In-Class Limits Strategy

This case study tests SDG with G1 placing synthetic cases in the class limits of the real data class prototyped from.

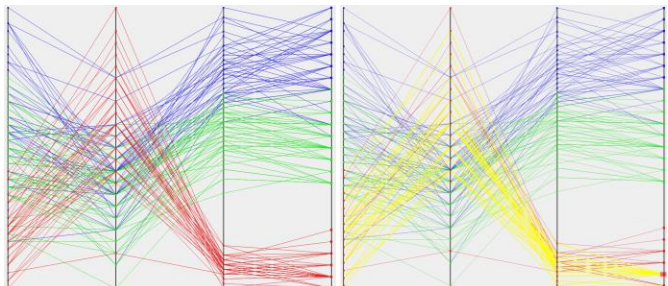

(a) Default Iris data in PC.   (b) 30 cases to be removed.
Figure 12. Iris data in parallel Coordinates.

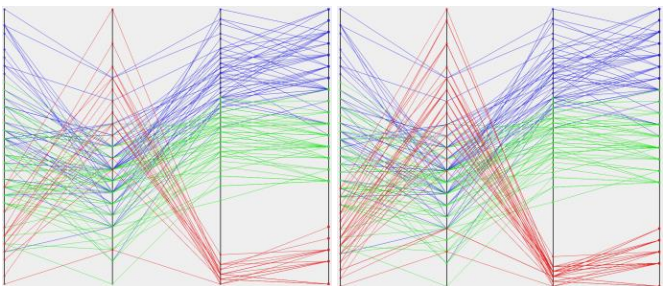

(a) Imbalanced data with 20 Setosa (red) cases, and 50 per other class.   (b) Data where 30 Setosa (red) cases are synthetic data.

Figure 13. Balancing test of Iris data with synthetic cases placed in limits of original real class.

As expected, the actual results are approximately the same with the synthetic data as with the original real data. Figures 12, 13 show balancing test process visualized and results with Table 3 show the comparison for Figure 13 of case study 1. We deleted 30 Setosa cases, then rebalanced the data with synthetic cases.

Table 4. Iris results without 30 deleted cases.

```
=============================================================================
Model Performance over 1000 independent cycles with 10-Fold Cross-Validation
Training Dataset: C:/Users/Alice/Documents/GitHub/DCVis/datasets/experiments/test2.csv
Exploration Dataset: C:/Users/Alice/Documents/GitHub/DCVis/datasets/fisher_iris.csv
=============================================================================
Model   CV Mean Acc.   CV STD of Acc.  Exp. Mean Acc.  Exp. STD of Acc. Best AUC  Worst AUC
DT       0.94           0.01            1.00            0.00             1.00      1.00
RF       0.94           0.01            1.00            0.00             1.00      1.00
XT       0.94           0.00            1.00            0.00             1.00      1.00
KNN      0.97           0.00            0.97            0.00             1.00      1.00
SVM      0.96           0.00            0.97            0.00             1.00      1.00
LDA      0.97           0.00            0.98            0.00             1.00      1.00
LR       0.96           0.00            0.97            0.00             1.00      1.00
Ridge    0.82           0.00            0.87            0.00             0.96      0.96
NB       0.94           0.00            0.96            0.00             0.99      0.99
MLP      0.97           0.01            0.98            0.02             1.00      0.83
SGD      0.79           0.05            0.83            0.12             0.99      0.98
GB       0.94           0.00            1.00            0.00             1.00      1.00
AB       0.92           0.00            0.96            0.00             1.00      1.00
XB       0.93           0.00            1.00            0.00             1.00      1.00
=============================================================================
```

Table 5. Results for 30 synthetic cases placed in bounds.

```
=============================================================================
Model Performance over 1000 independent cycles with 10-Fold Cross-Validation
Training Dataset: C:/Users/Alice/Documents/GitHub/DCVis/datasets/experiments/test10.csv
Exploration Dataset: C:/Users/Alice/Documents/GitHub/DCVis/datasets/fisher_iris.csv
=============================================================================
Model   CV Mean Acc.   CV STD of Acc.  Exp. Mean Acc.  Exp. STD of Acc. Best AUC  Worst AUC
DT       0.94           0.01            1.00            0.00             1.00      1.00
RF       0.95           0.00            1.00            0.00             1.00      1.00
XT       0.95           0.00            1.00            0.00             1.00      1.00
KNN      0.95           0.00            0.97            0.00             1.00      1.00
SVM      0.97           0.00            0.97            0.00             1.00      1.00
LDA      0.97           0.00            0.98            0.00             1.00      1.00
LR       0.97           0.00            0.97            0.00             1.00      1.00
Ridge    0.86           0.00            0.87            0.00             0.96      0.96
NB       0.95           0.00            0.96            0.00             0.99      0.99
MLP      0.97           0.01            0.98            0.01             1.00      1.00
SGD      0.83           0.04            0.85            0.11             0.99      0.98
GB       0.95           0.00            1.00            0.00             1.00      1.00
AB       0.93           0.00            0.96            0.00             1.00      1.00
XB       0.94           0.00            1.00            0.00             1.00      1.00
=============================================================================
```

## D. Case Study IV: Out-of-Class Limits Strategy

In this case study, we place generated synthetic cases outside of the class limits of the real data class. Figures 14 shows two placement versions in PC on coordinate $X_3$.

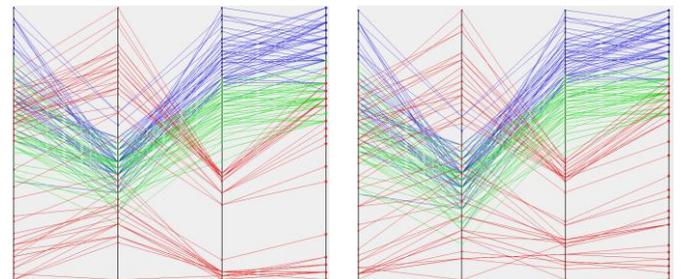

(a) Case Study IV Version 1   (b) Case Study IV Version 2
Figure 14. Iris with synthetic red (Setosa) cases placed out-of-limits in PC.

Table 6. Results of worsened performance in two tests, first experiment.

```
=============================================================================
Model Performance over 1000 independent cycles with 10-Fold Cross-Validation
Training Dataset: C:/Users/Alice/Documents/GitHub/DCVis/datasets/experiments/test13.csv
Exploration Dataset: C:/Users/Alice/Documents/GitHub/DCVis/datasets/fisher_iris.csv
=============================================================================
Model   CV Mean Acc.   CV STD of Acc.  Exp. Mean Acc.  Exp. STD of Acc. Best AUC  Worst AUC
DT       0.93           0.01            1.00            0.00             1.00      1.00
RF       0.94           0.00            1.00            0.00             1.00      1.00
XT       0.95           0.00            1.00            0.00             1.00      1.00
KNN      0.95           0.00            0.97            0.00             1.00      1.00
SVM      0.97           0.00            0.97            0.00             1.00      1.00
LDA      0.96           0.00            0.97            0.00             1.00      1.00
LR       0.96           0.00            0.97            0.00             1.00      1.00
Ridge    0.77           0.00            0.81            0.00             1.00      1.00
NB       0.93           0.00            0.95            0.00             0.99      0.99
MLP      0.97           0.01            0.98            0.01             1.00      1.00
SGD      0.82           0.04            0.84            0.12             1.00      0.99
GB       0.95           0.00            1.00            0.00             1.00      1.00
AB       0.87           0.00            0.85            0.00             0.99      0.99
XB       0.93           0.00            1.00            0.00             1.00      1.00
=============================================================================
```

Table 7. Results of worsened performance in two tests, second experiment.

Putting cases outside of the original bounds of the class resulted in lower accuracies relative to the results of case studies II and III, but not significantly deteriorated. While there is a drop in the accuracies, this is to be expected, since the synthetic data generated are outside of the bounds of the red (Setosa) class in $X_3$, but not in overlap areas of other classes as Figure 14 shows.

Table 7. Results of worsened performance in two tests, second experiment.

```
=========================================================================
Model Performance over 1000 independent cycles with 10-Fold Cross-Validation
Training Dataset: C:/Users/Alice/Documents/GitHub/DCVis/datasets/experiments/test14.csv
Exploration Dataset: C:/Users/Alice/Documents/GitHub/DCVis/datasets/fisher_iris.csv
=========================================================================
Model  CV Mean Acc.  CV STD of Acc.  Exp. Mean Acc.  Exp. STD of Acc. Best AUC  Worst AUC
DT     0.93          0.01            1.00            0.00             1.00      1.00
RF     0.94          0.00            1.00            0.00             1.00      1.00
XT     0.95          0.00            1.00            0.00             1.00      1.00
KNN    0.95          0.00            0.97            0.00             1.00      1.00
SVM    0.97          0.00            0.97            0.00             1.00      1.00
LDA    0.95          0.00            0.96            0.00             1.00      1.00
LR     0.96          0.00            0.97            0.00             1.00      1.00
Ridge  0.79          0.00            0.81            0.00             0.99      0.99
NB     0.93          0.00            0.95            0.00             0.99      0.99
MLP    0.97          0.01            0.98            0.00             1.00      1.00
SGD    0.82          0.04            0.84            0.12             1.00      0.99
GB     0.95          0.00            1.00            0.00             1.00      1.00
AB     0.92          0.00            0.96            0.00             1.00      1.00
XB     0.93          0.00            1.00            0.00             1.00      1.00
=========================================================================
```

### E. Case Study V: Proportional Randomization Strategy

Here we generated synthetic cases by placing them in random locations: (1) proportionally, or (2) without regard to proportionality or n-D distributions. Results for (1) and (2) are shown in Figures 15, 16, and Figures 17, 18, respectively with mostly improved accuracy in (1), degraded in (2).

Tables 8, 9 show the results of the proportionally randomized case study tests repeated for verification of randomization capabilities. This had positive performance for Iris data due to the n-D distribution consistency between real and synthetic data.

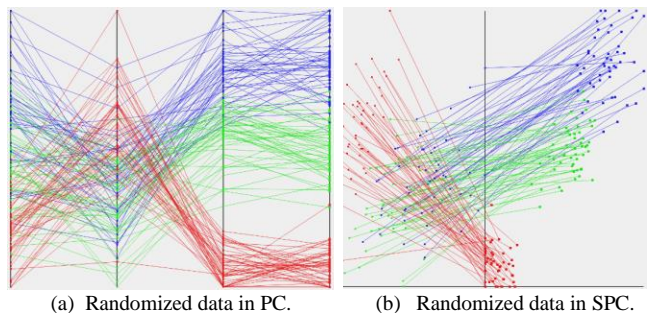

(a) Randomized data in PC. (b) Randomized data in SPC.

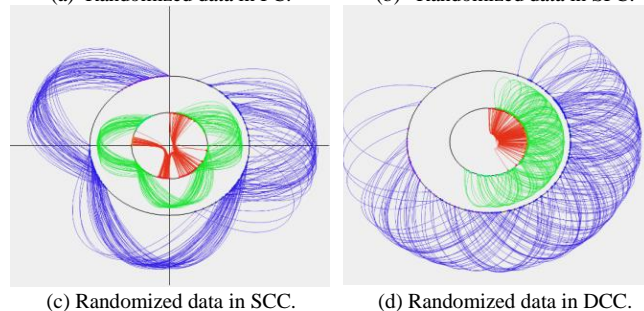

(c) Randomized data in SCC. (d) Randomized data in DCC.
Figure 16: Proportionally Randomized Iris data in GLCs, second test.

Table 8. Proportional randomization evaluation, first test from figure 15.

```
=========================================================================
Model Performance over 1000 independent cycles with 10-Fold Cross-Validation
Training Dataset: C:/Users/Alice/Documents/GitHub/DCVis/datasets/experiments/iris_inbounds1.csv
Exploration Dataset: C:/Users/Alice/Documents/GitHub/DCVis/datasets/fisher_iris.csv
=========================================================================
Model  CV Mean Acc.  CV STD of Acc.  Exp. Mean Acc.  Exp. STD of Acc. Best AUC  Worst AUC
DT     0.95          0.01            0.97            0.00             0.98      0.98
RF     0.94          0.01            0.97            0.01             1.00      1.00
XT     0.93          0.00            0.98            0.01             1.00      1.00
KNN    0.96          0.00            0.97            0.00             1.00      1.00
SVM    0.95          0.00            0.97            0.00             1.00      1.00
LDA    0.96          0.00            0.98            0.00             1.00      1.00
LR     0.95          0.00            0.98            0.00             1.00      1.00
Ridge  0.85          0.00            0.83            0.00             0.96      0.96
NB     0.91          0.00            0.96            0.00             0.99      0.99
MLP    0.94          0.01            0.98            0.01             1.00      1.00
SGD    0.83          0.04            0.84            0.10             0.99      0.98
GB     0.92          0.00            0.97            0.00             1.00      1.00
AB     0.92          0.00            0.95            0.00             1.00      1.00
XB     0.93          0.00            0.97            0.00             1.00      1.00
=========================================================================
```

Table 9. Proportional randomization evaluation, second test from figure 16.

```
=========================================================================
Model Performance over 1000 independent cycles with 10-Fold Cross-Validation
Training Dataset: C:/Users/Alice/Documents/GitHub/DCVis/datasets/experiments/iris_inbounds2.csv
Exploration Dataset: C:/Users/Alice/Documents/GitHub/DCVis/datasets/fisher_iris.csv
=========================================================================
Model  CV Mean Acc.  CV STD of Acc.  Exp. Mean Acc.  Exp. STD of Acc. Best AUC  Worst AUC
DT     0.94          0.01            0.96            0.01             0.97      0.96
RF     0.95          0.00            0.96            0.00             1.00      1.00
XT     0.95          0.01            0.96            0.00             1.00      1.00
KNN    0.96          0.00            0.97            0.00             1.00      1.00
SVM    0.95          0.00            0.95            0.00             1.00      1.00
LDA    0.97          0.00            0.95            0.00             1.00      1.00
LR     0.96          0.00            0.96            0.00             1.00      1.00
Ridge  0.83          0.00            0.86            0.00             0.96      0.96
NB     0.97          0.00            0.96            0.00             0.99      0.99
MLP    0.96          0.01            0.98            0.00             1.00      1.00
SGD    0.84          0.04            0.86            0.10             0.99      0.98
GB     0.94          0.00            0.94            0.00             1.00      0.99
AB     0.91          0.00            0.93            0.00             1.00      1.00
XB     0.93          0.00            0.97            0.00             1.00      1.00
=========================================================================
```

Figure 17 shows visualizes the results of the out-of-limits out-of-class-limits case study tests repeated twice for verification of randomization strategy capabilities with Table 10 and 11 shows the results of this case study.

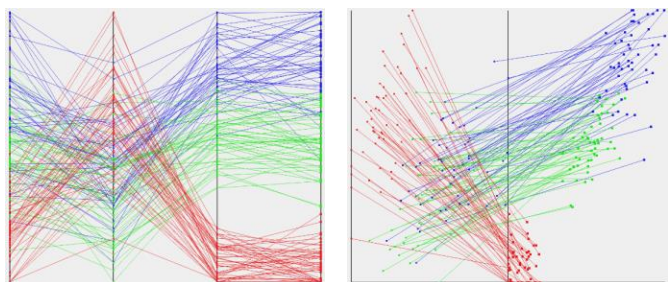

(a) Randomized data in PC. (b) Randomized data in SPC.

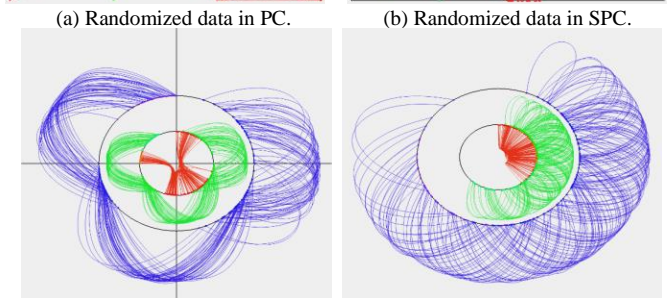

(c) Randomized data in SCC. (d) Randomized data in SCC.
Figure 15: Proportionally Randomized Iris data in GLCs, first test.

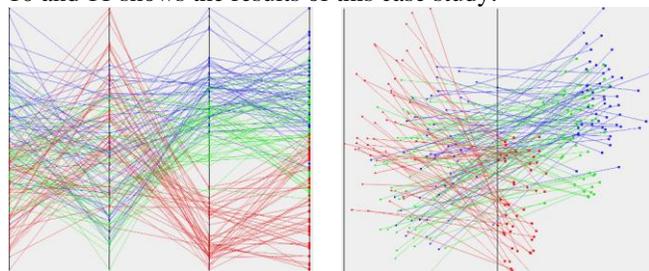

(a) Randomized Iris data in PC. (b) Randomized Iris data in SPC.

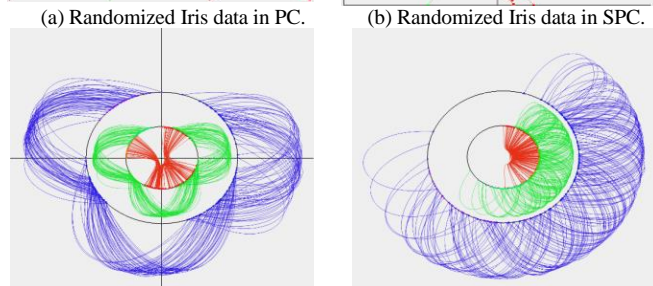

(c) Randomized Iris data in SCC. (d) Randomized Iris data in DCC.
Figure 17. Unporportionaly Randomized Iris data in GLCs.

Table 10. Performance worsened for unproportional randomization, test one.

```
=============================================================================
Model Performance over 1000 independent cycles with 10-Fold Cross-Validation
Training Dataset: C:/Users/Alice/Documents/GitHub/DCVis/datasets/experiments/fisher_nobounds1.csv
Exploration Dataset: C:/Users/Alice/Documents/GitHub/DCVis/datasets/fisher_iris.csv
=============================================================================
Model   CV Mean Acc.   CV STD of Acc.   Exp. Mean Acc.   Exp. STD of Acc.   Best AUC   Worst AUC
DT      0.75           0.01             0.85             0.01               0.90       0.88
RF      0.84           0.01             0.85             0.02               0.98       0.97
XT      0.83           0.01             0.85             0.01               0.98       0.96
KNN     0.85           0.00             0.92             0.00               0.97       0.97
SVM     0.87           0.00             0.93             0.00               0.99       0.99
LDA     0.89           0.00             0.95             0.00               0.99       0.99
LR      0.86           0.00             0.95             0.00               0.99       0.99
Ridge   0.78           0.00             0.83             0.00               0.96       0.96
NB      0.87           0.00             0.91             0.00               0.99       0.99
MLP     0.79           0.01             0.96             0.01               1.00       1.00
SGD     0.72           0.03             0.78             0.09               1.00       0.97
GB      0.82           0.00             0.89             0.01               0.98       0.98
AB      0.78           0.00             0.93             0.00               0.97       0.97
XB      0.81           0.00             0.87             0.00               0.97       0.97
=============================================================================
```

Table 11. Performance worsened for unproportional randomization, test two.

```
=============================================================================
Model Performance over 1000 independent cycles with 10-Fold Cross-Validation
Training Dataset: C:/Users/Alice/Documents/GitHub/DCVis/datasets/experiments/fisher_nobounds2.csv
Exploration Dataset: C:/Users/Alice/Documents/GitHub/DCVis/datasets/fisher_iris.csv
=============================================================================
Model   CV Mean Acc.   CV STD of Acc.   Exp. Mean Acc.   Exp. STD of Acc.   Best AUC   Worst AUC
DT      0.83           0.01             0.81             0.00               0.86       0.86
RF      0.84           0.01             0.86             0.01               0.99       0.97
XT      0.84           0.01             0.88             0.01               1.00       0.98
KNN     0.82           0.00             0.87             0.00               0.97       0.97
SVM     0.84           0.00             0.96             0.00               1.00       1.00
LDA     0.85           0.00             0.97             0.00               1.00       1.00
LR      0.85           0.00             0.96             0.00               0.99       0.99
Ridge   0.73           0.00             0.79             0.00               0.99       0.99
NB      0.85           0.00             0.94             0.00               0.99       0.99
MLP     0.83           0.01             0.96             0.01               1.00       1.00
SGD     0.73           0.03             0.79             0.11               1.00       0.99
GB      0.85           0.00             0.85             0.00               0.98       0.98
AB      0.78           0.00             0.89             0.00               0.95       0.95
XB      0.85           0.00             0.86             0.00               0.98       0.98
=============================================================================
```

## IV. DCVIS TOOL FUNCTIONALITY

DCVis, the Dynamic Coordinates Visualization System, was originally built from DSCVis, the Dynamic Scaffold Coordinates Visualization System [11]. DCVis visualizes n-D data in interchangeable GLCs. The software supports GLCs of PC, SPC, Dynamic Scaffold Coordinates 1 & 2 (DSC1 & DSC2), SCC, and DCC. DCVis visualizes tabular data loaded from a .txt or .csv file containing ML data where one column contains labels with the the header of 'class'. Use of GLCs benefits the SDG-ADL aglorithm by accentuating different data properties by each GLC, enabling enhanced discoverability of classifiable features. Visualized properties can be tuned in any GLC with attributes and class orders or by inverting attributes, toggling the ranges of [0,1] and [1, 0]. SCC allows conflicting points to be replotted for further classification, DCC plotted function coefficients can be scaled, defaults by LDA.

Options for visuals are in the UI for attribute transparency, background, axes, and class colors, with visibility of lines, points, and CC sectors. Clipping regions can be drawn to select cases with the mouse, left-click selects directly under the mouse, two right-clicks draws a selection region, middle-click grows the selection region, selected cases are highlighted. Q, E keyboard keys select the previous and next cases in the dataset, W, A move the selected cases up and down in real-time, C clones them, and D deletes them. Selecting cases allows for analyzing for purity with UI buttons or to form associative classification rules, which can be chain together outlined areas. Rule assignment sets colors regions by the class color if pure or off-white otherwise. A trace mode button shows n-D cases in unique colors.Transformed models can be saved into a .csv file for visualization or classifier testing.

DCVis is freely available for personal and commercial use under the MIT license at github.com/CWU-VKD-LAB/DCVis. How to use the software is documented in the project readme.

## V. CONCLUSIONS

Synthetic n-D data for ML/AI is of critical importance, however, SDG must be evaluated for assurance of quality synthetic data and resultant models trained on that synthetic data. Therefore, we proposed the unified SDG-ADL algorithm combining tasks of data generation and labeling in an automatable process with the tool of GLCs. We demonstrated this approach, and it shows that SDG can improve resultant ML/AI models. Future work should explore scaling to larger dimensionality of data including applied use cases such as using SDG for recommendation system tuning with user preferences.

## VI. ACKNOWLEDGMENTS